\documentclass[11pt]{article}

\usepackage[margin=1in]{geometry}
\usepackage[T1]{fontenc}
\usepackage[utf8]{inputenc}
\usepackage{lmodern}
\usepackage{amsmath,amssymb}
\usepackage{graphicx}
\usepackage{multirow}
\usepackage{bm}
\usepackage{nicefrac}
\usepackage{url}
\usepackage[hidelinks]{hyperref}
\usepackage[numbers,sort&compress]{natbib}
\graphicspath{{figures/}}

\newcommand{\affmark}[1]{\textsuperscript{#1}}

\title{Machine Learning--Augmented Acceleration of Iterative Ptychographic Reconstruction}

\author{%
Bowen Zheng\affmark{1},
Katayun Kamdin\affmark{1},
David Shapiro\affmark{1},
Alexander Ditter\affmark{1},
Dayne Sasaki\affmark{1,2}, \\
Emma Bernard\affmark{3},
Roopali Kukreja\affmark{3},
Sheena K.K. Patel\affmark{4},
Petrus H. Zwart\affmark{5,6,7},\\
Slavom\'ir Nem\v{s}\'ak\affmark{1,8},
Apurva Mehta\affmark{9},
Nicholas Schwarz\affmark{10},
Alexander Hexemer\affmark{1},
Tanny Chavez\affmark{1,*}\\[0.75em]
\begin{minipage}{1.0\textwidth}
\centering
\small
\affmark{1}Advanced Light Source, Lawrence Berkeley National Laboratory, Berkeley, CA 94720, USA\\
\affmark{2}Department of Physics, Massachusetts Institute of Technology, Cambridge, MA 02139, USA\\
\affmark{3}Department of Materials Science and Engineering, University of California Davis, Davis, CA 95616, USA\\
\affmark{4}Center for Memory and Recording Research, University of California San Diego, La Jolla, CA 92093, USA.\\
\affmark{5}Center for Advanced Mathematics for Energy Research Applications, Lawrence Berkeley National Laboratory, Berkeley, CA 94720, USA\\
\affmark{6}Molecular Biophysics and Integrated Bioimaging Division, Lawrence Berkeley National Laboratory, Berkeley, CA 94720, USA\\
\affmark{7}Berkeley Synchrotron Infrared Structural Biology Program, Lawrence Berkeley National Laboratory, Berkeley, CA 94720, USA\\
\affmark{8}Department of Physics and Astronomy, University of California, Davis, Davis, CA 95616, USA\\
\affmark{9}Linac Coherent Light Source (LCLS), SLAC National Accelerator Laboratory, Menlo Park, CA 94205, USA\\
\affmark{10}Argonne National Laboratory, 9700 South Cass Avenue, Lemont, IL 60439, USA\\[0.5em]
\affmark{*}\texttt{tanchavez@lbl.gov}
\end{minipage}
}

\date{}

\begin{document}
\maketitle

\begin{abstract}
Iterative ptychographic reconstruction algorithms are widely used for coherent diffractive imaging but can exhibit slow convergence under realistic experimental conditions. We propose a machine learning--augmented approach that accelerates iterative ptychographic reconstruction by introducing a learned fast-forward operator applied during reconstruction. Following an initial warm-up using standard iterations, the fast-forward operator advances the reconstruction toward a more converged state, after which conventional iterative updates are resumed. This strategy preserves the physical consistency and flexibility of established ptychographic solvers while reducing the number of iterations required for convergence. The model is trained on diverse ptychographic datasets and evaluated on experimental data acquired in different years, demonstrating robustness and temporal generalization. Compared with conventional iterative solvers, the machine learning-augmented method consistently achieves comparable or improved reconstruction quality while converging faster in terms of Poisson negative log-likelihood, yielding a near two-fold reduction in wall-clock time. The approach has been integrated into an existing reconstruction pipeline and deployed in production at a synchrotron beamline, demonstrating practicality for real-time experimental operation.

\end{abstract}

%%%%%%%%%%%%%%%%%%%%%%%%%%  body  %%%%%%%%%%%%%%%%%%%%%%%%%%

\section{Introduction}
Modern synchrotron and electron microscopy facilities are undergoing transformative upgrades in detector technology and beam brightness. Fourth-generation synchrotron sources now deliver coherent X-ray flux orders of magnitude higher than their predecessors, while next-generation detectors capture diffraction patterns at kilohertz frame rates~\cite{rodenburg2004phase, miao2025computational}. For ptychography, a widely used computational imaging technique that recovers high-resolution complex-valued object information from intensity only diffraction measurements, this dramatic increase in data acquisition rates has created a critical computational bottleneck. Traditional iterative reconstruction algorithms can no longer provide the near-real-time feedback required to guide experimental decision-making, transforming what was once a computational challenge into an operational constraint that limits scientific throughput.

The core challenge lies in preserving reconstruction fidelity while achieving the acceleration needed for high-throughput operation. Classical ptychographic reconstruction relies on iterative projection or optimization-based algorithms such as the extended ptychographic iterative engine (ePIE)~\cite{maiden2009improved, maiden2017further}, the difference map~\cite{thibault2008high, thibault2009probe}, and maximum-likelihood methods~\cite{thibault2012maximum}. These algorithms recover missing phase information by enforcing consistency between measured diffraction intensities and spatial overlap constraints through gradient-based minimization of a well-defined objective function. This gradient-based optimization provides crucial guarantees: convergence to physically consistent solutions, interpretable intermediate states, and robustness to experimental imperfections such as scan position errors and partial coherence~\cite{miao2025computational}. However, achieving these guarantees comes at substantial computational cost, with reconstructions often requiring hundreds of iterations and minutes to hours of processing time per dataset.

Machine learning (ML) has emerged as a potential solution for accelerating ptychographic reconstruction. Many current ML methods rely on single-shot direct mappings from diffraction patterns to reconstructed objects~\cite{cherukara2020ai, babu2023deep, lee2025deep, vu2025pid3net, chang2023deep, hoidn2023physics}, treating reconstruction as a regression problem. While these approaches enable rapid inference once trained, they fundamentally replace the physics-based optimization with a black-box predictor. This replacement eliminates the gradient-based refinement mechanism that ensures physical consistency with measured data, provides interpretability through well-defined objective functions, and enables adaptive handling of experimental variations. Furthermore, these methods typically train on synthetic or simulated data rather than experimental measurements and assume fixed or known probe configurations, raising concerns about robustness in operational beamline environments where experimental conditions vary continuously. Plug-and-play (PnP) priors~\cite{denker2025plug, sun2019regularized, aslan2021joint} have also been explored as a single-shot ML alternative, where the hand-craft, explicit regularizer (such as the $L_2$ norm) is replaced by a learned image denoiser. Instead of solving a proximal operator analytically, the denoiser processes the current image estimate at each iteration. Compared with the single-shot ML method, PnP priors preserve the physics-based iterations. However, they replace the cheap proximal step in classical algorithms with a more expensive denoiser call, which is a full neural network forward pass executed at every iteration. This could add substantial computational overhead to the classical iterative algorithm.

To achieve ptychographic reconstruction that is both efficient and physically valid, we introduce a single-step ML-augmented acceleration strategy that preserves the gradient-based minimization framework while achieving substantial speedup. Rather than replacing the iterative solver, our approach inserts a learned fast-forward operator that advances early-stage reconstructions toward more converged states. Critically, this operator acts \emph{within} the optimization loop \emph{only once}: conventional gradient-based iterations perform an initial warm-up, the ML operator provides a single acceleration step, and gradient-based optimization then resumes to drive the solution toward physical consistency with measured data. This hybrid design maintains all guarantees of the underlying solver, convergence to statistically optimal reconstructions, interpretable loss landscapes, and robustness to experimental imperfections, while reducing the total iteration count required for convergence. The method is trained solely on historical experimental ptychographic datasets from multiple years and sample types, and evaluated on independently acquired data from different years to rigorously assess cross-year generalization. We demonstrate comparable or improved reconstruction quality and accelerated convergence on the test data, and report successful deployment in production at an operational synchrotron beamline, where the approach routinely provides a near two-fold reduction in reconstruction time while maintaining or improving reconstruction quality equivalent to conventional iterative solvers.

% --------------------
\section{Method}
\subsection{Overall Framework}
\begin{figure}[!ht]
    \centering
    \includegraphics[width=0.99\linewidth]{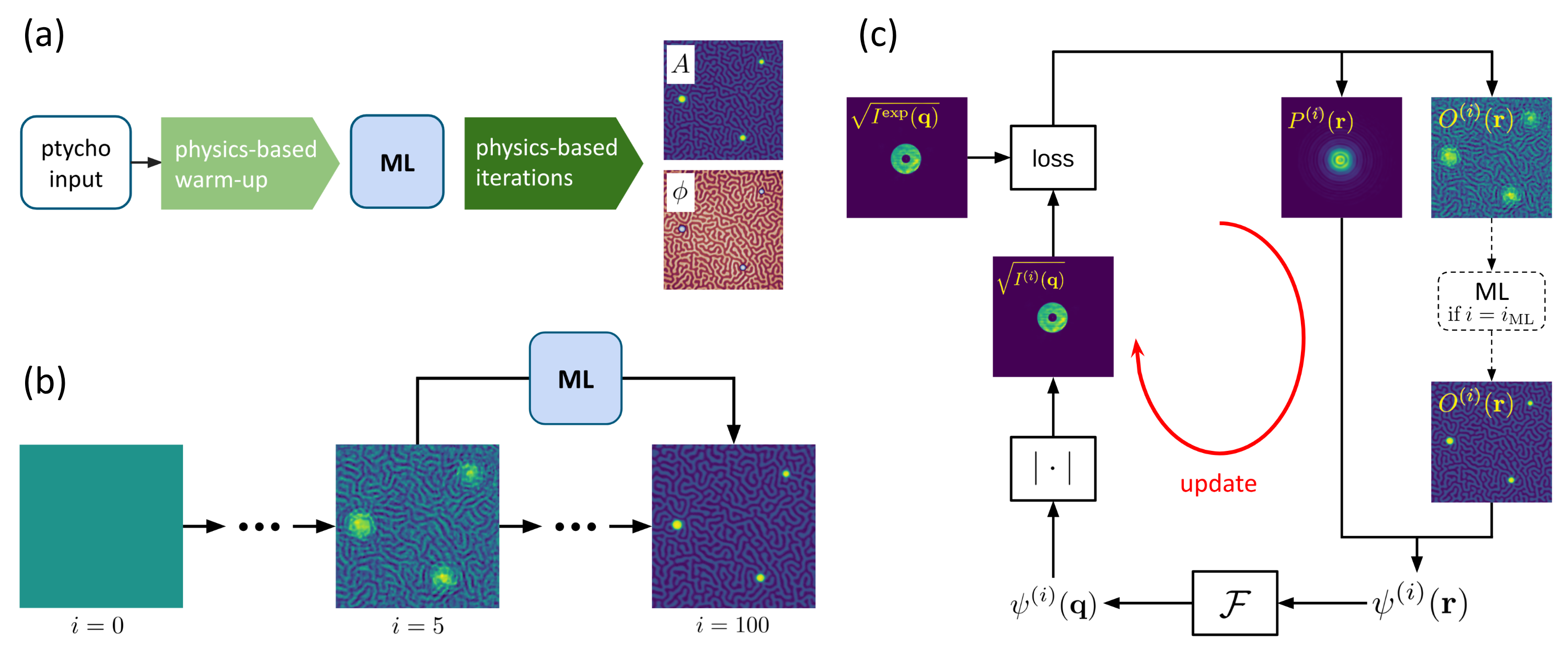}
    \caption{Schematic of the proposed ptychographic reconstruction framework. (a) High-level workflow. Ptychographic reconstruction inputs are first provided to the physics-based iterative solver. A small number of ptychographic iterations are performed as a warm-up stage. The resulting complex-valued object estimate is then passed to an ML-based fast-forward operator, which predicts an updated object. Subsequent ptychographic iterations are carried out using this prediction as the new initialization. (b) ML fast-forward operator objective. The goal is to advance an early reconstruction state toward a more converged solution. (c) Algorithmic overview of the ptychographic update process within a single iteration. Probe and object estimates are iteratively refined by enforcing consistency between measured and reconstructed diffraction amplitudes. At the beginning of iteration $i_{\mathrm{ML}}$, the ML-based fast-forward operator updates the full object. In (a--c) as well as Fig.~\ref{fig:3}, a cobalt-based magnetic material is used as an illustrative example and is included in the training data.}

    \label{fig:1}
\end{figure}

The proposed framework augments an existing iterative ptychographic reconstruction pipeline with an ML-based fast-forward operator, without replacing the underlying physics-based solver. Instead, the learned model is inserted as an auxiliary module that acts on intermediate object estimates generated during reconstruction. As shown in Fig.~\ref{fig:1}(a), standard ptychographic inputs, including measured diffraction patterns and their associated scan positions, are first processed by a conventional reconstruction engine, which performs a small number of iterations as a warm-up stage. The resulting complex-valued object estimate is then passed to the ML-based fast-forward operator, which predicts an improved estimate closer to convergence. This prediction is used to reinitialize the iterative solver, allowing subsequent ptychographic iterations to resume from a more advanced state. The learning objective of the fast-forward operator, illustrated in Fig.~\ref{fig:1}(b), is to advance an early-stage object reconstruction toward a more converged solution, effectively reducing the number of solver iterations required by performing a single inference step. Importantly, the ML inference is performed only once during the entire reconstruction process, resulting in minimal additional computational overhead compared to the overall cost of iterative ptychographic reconstruction. This hybrid design combines the robustness physics-based reconstruction with the efficiency gains of data-driven acceleration. 

In this work, the method is implemented within the CDTools ptychographic reconstruction framework~\cite{Levitan_CDTools}, which provides iterative solvers based on automatic differentiation. 
Fig.~\ref{fig:1}(c) illustrates the algorithmic workflow of the proposed ML-augmented reconstruction. During ptychographic iteration $i$, the probe $P^{(i)}(\mathbf{r})$ interacts with an object patch $O^{(i)}(\mathbf{r}-\mathbf{R}_k)$ at scan position $\mathbf{R}_k$ to form the real-space exit wave
\begin{equation}
\psi_k^{(i)}(\mathbf{r}) 
= P^{(i)}(\mathbf{r}) \, O^{(i)}(\mathbf{r}-\mathbf{R}_k),
\end{equation}
where $k$ indexes the scan position. The exit wave is propagated to reciprocal space using the Fourier transform $\mathcal{F}$,
\begin{equation}
\psi_k^{(i)}(\mathbf{q}) 
= \mathcal{F} \left\{ \psi_k^{(i)}(\mathbf{r}) \right\},
\end{equation}
from which the predicted diffraction intensity is computed as
\begin{equation}
I_k^{(i)}(\mathbf{q}) 
= \left| \psi_k^{(i)}(\mathbf{q}) \right|^2.
\end{equation}
The predicted amplitudes are compared with the experimentally measured amplitudes using a mean-squared-error (MSE) loss,
\begin{equation}
\mathcal{L}^{(i)} 
= \frac{1}{K} \sum_{k=1}^{K}
\left\|
\sqrt{I_k^{(i)}(\mathbf{q})}
-
\sqrt{I_k^{\mathrm{exp}}(\mathbf{q})}
\right\|_2^2,
\end{equation}
where $\left\| \cdot \right\|_2$ denotes the $L_2$ norm; $K$ denotes the number of diffraction patterns in the minibatch. The loss is backpropagated through the forward model to compute gradients with respect to both the probe and object estimates,
\begin{equation}
P^{(i+1)} 
= P^{(i)} 
- \eta \, \frac{\partial \mathcal{L}^{(i)}}{\partial P^{(i)}},
\qquad
O^{(i+1)} 
= O^{(i)} 
- \eta \, \frac{\partial \mathcal{L}^{(i)}}{\partial O^{(i)}},
\end{equation}
where $\eta$ denotes the learning rate. The reconstruction is carried out in minibatches to reduce memory usage and enable efficient optimization on large datasets. At the beginning of iteration $i = i_{\mathrm{ML}}$, an ML-based fast-forward operator $\mathcal{G}_\theta$ is applied once to the full object estimate,
\begin{equation}
O^{'(i_{\mathrm{ML}})} 
\leftarrow 
\mathcal{G}_\theta \big( O^{(i_{\mathrm{ML}})} \big),
\end{equation}
to accelerate convergence toward a physically consistent solution.

\subsection{Network Architecture}
\begin{figure}[!ht]
    \centering
    \includegraphics[width=0.99\linewidth]{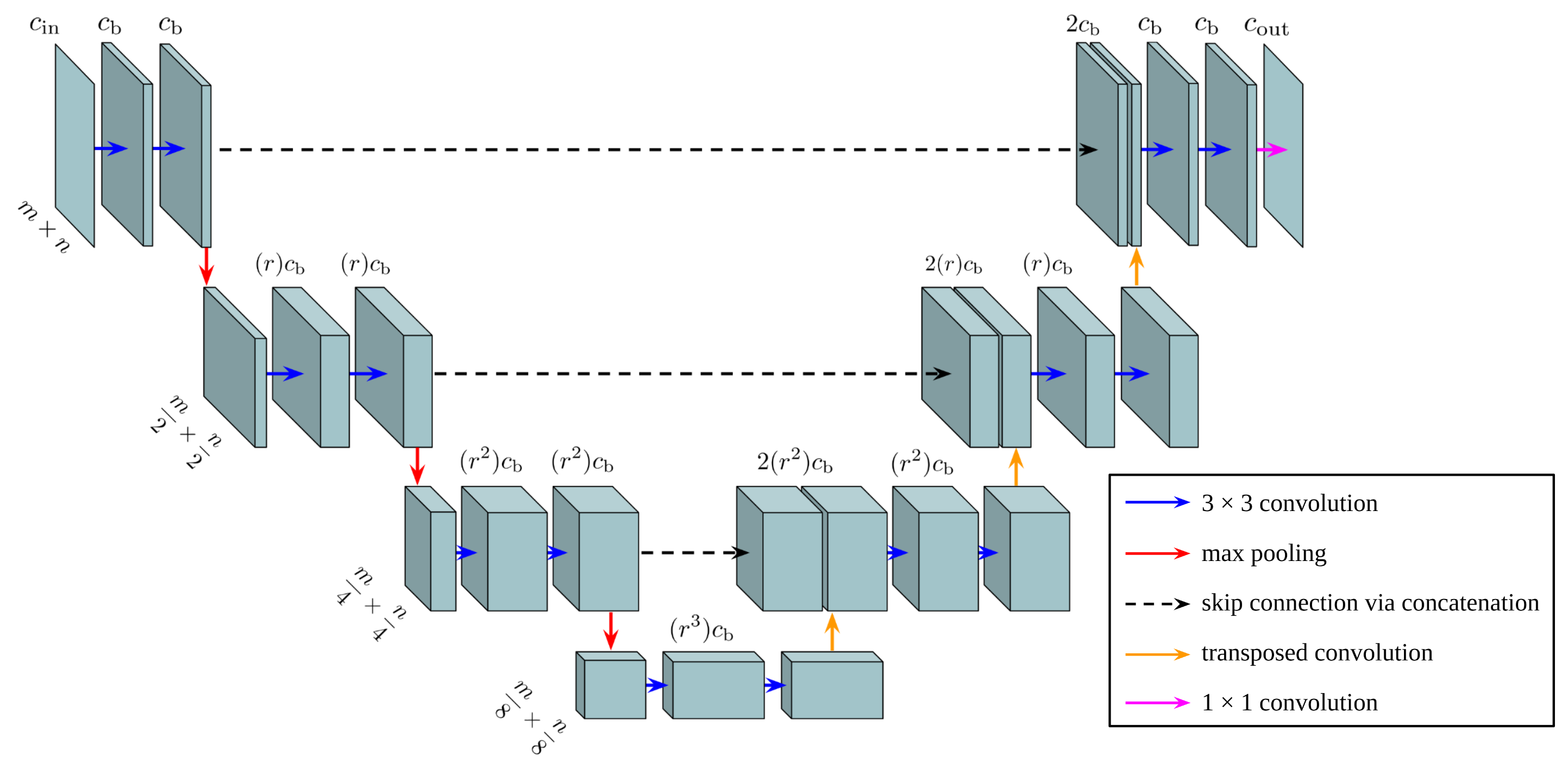}
    \caption{
    Schematic of the ML model architecture. The ML-based fast-forward operator is implemented as a U-Net with $3\times3$ convolutions and max pooling in the encoder, transposed convolutions in the decoder, and skip connections via feature-map concatenation. A final $1\times1$ convolution produces the output object.
    }

    \label{fig:2}
\end{figure}

Fig.~\ref{fig:2} shows the ML model architecture used for the fast-forward operator. The operator is implemented as a U-Net~\cite{ronneberger2015u} consisting of an encoder with $3\times3$ convolutions and max-pooling layers, and a decoder with transposed convolutions for upsampling. Skip connections are introduced via feature-map concatenation between corresponding encoder and decoder layers to preserve spatial detail. A final $1\times1$ convolution produces the output object. In Fig.~\ref{fig:2}, $m$ and $n$ denote the input image dimensions, $c_{\mathrm{in}}$ and $c_{\mathrm{out}}$ denote the number of input and output channels, respectively, $c_b$ denotes the number of base channels, and $r$ denotes the channel growth rate across network depths. Because standard convolutional layers operate on real-valued tensors, each complex-valued object field is represented using two real-valued channels corresponding to its real and imaginary components, resulting in $c_{\mathrm{in}} = c_{\mathrm{out}} = 2$. In this work, U-Net is implemented using the \texttt{dlsia} library~\cite{roberts2024dlsia}, which allows easy tuning of architecture-governing parameters. A key advantage of U-Net, as a fully convolutional architecture, is its ability to process input images of arbitrary spatial resolution without architectural modification or retraining. This property is shared by fully convolutional models and contrasts with standard Vision Transformer~\cite{dosovitskiy2020image} architectures, which typically assume fixed input sizes due to patch embeddings and positional encodings, though variants exist to relax this constraint. Additionally, convolutional models scale approximately linearly with image size, whereas global self-attention scales quadratically with the number of patches, leading to higher memory and runtime costs for large inputs. This flexibility is particularly important in our setting, where object images vary in spatial size.

\subsection{Training Strategy}
\begin{figure}[!ht]
    \centering
    \includegraphics[width=0.85\linewidth]{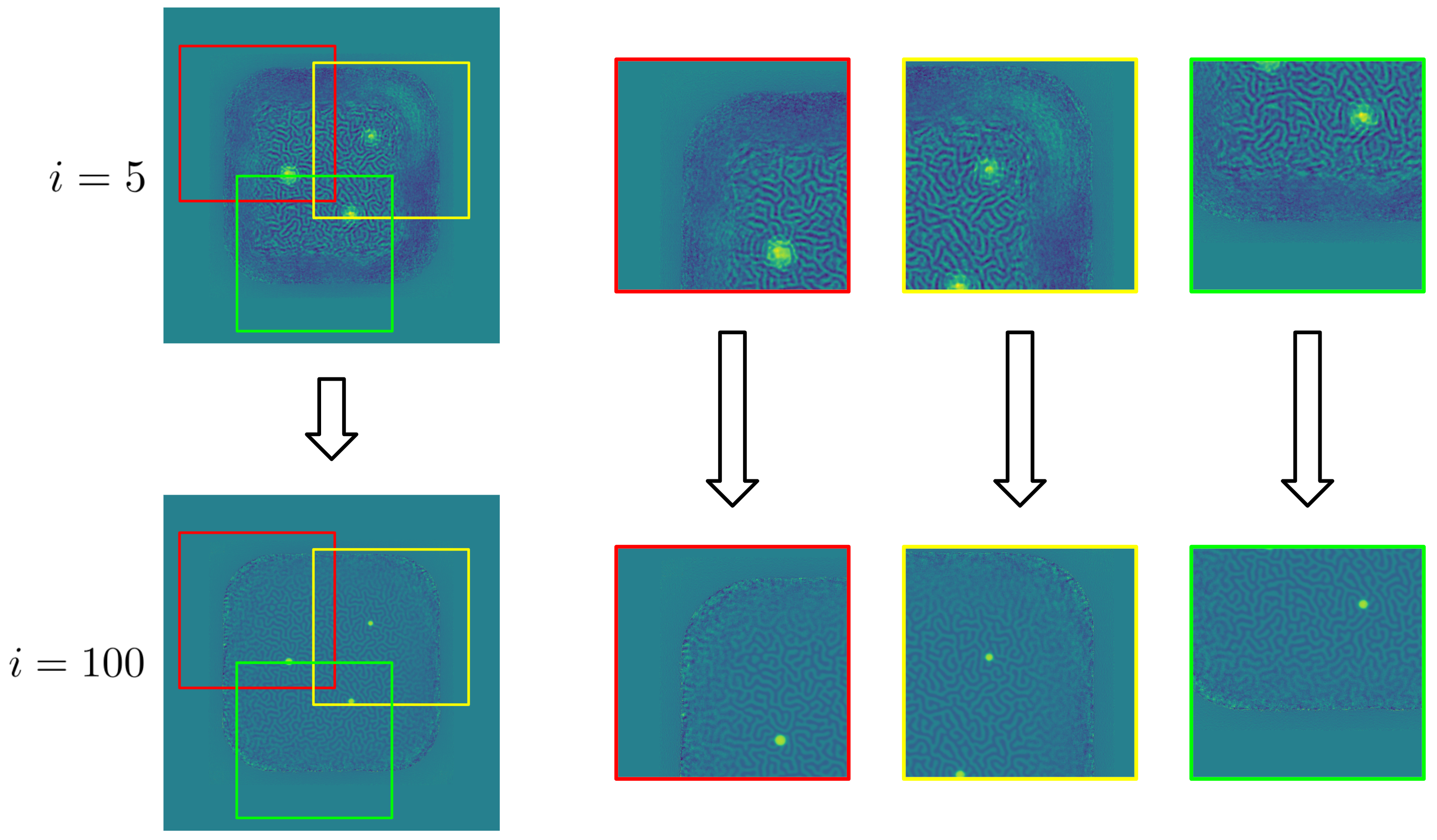}
    \caption{
    Illustration of the training data preparation strategy. Each complex-valued object is zero-padded, and $N$ random spatial patches are sampled from both the input object (top row, $i=5$) and the target object (bottom row, $i=100$). Patches extracted from the same spatial locations form corresponding input--target training pairs. For clarity, three representative patches are shown in this figure. 
    }

    \label{fig:3}
\end{figure}

The model is trained in a supervised manner using pairs of early- and later-stage ptychographic reconstructions. For each training dataset, a conventional CDTools reconstruction is first performed, and object estimates at iterations 5 and 100 are extracted. The object at iteration 5 is treated as the network input, representing an early-stage reconstruction, while the object at iteration 100 serves as the training target, representing a more converged solution. In practice, iteration 5 is not pure noise but still far from well reconstructed, whereas iteration 100 is typically well converged. This choice reflects a balance: mapping very noisy objects to well-reconstructed ones is challenging, while mapping moderately reconstructed states (e.g., iteration 20) to iteration 100 may offer limited learning benefit. To increase training diversity and reduce memory requirements, $N=8$ random spatial patches of 512$\times$512 pixels are sampled from both the input and target objects. Corresponding patches from the same spatial locations are used to form training pairs. The total number of training examples is therefore equal to the number of ptychographic datasets multiplied by $N$. The complex-valued objects used for training are zero-padded. This is important because inference data usually contains padded regions, and exposing the model to this during training helps it learn the correct behavior in those regions. The above data preparation strategy is illustrated in Fig.~\ref{fig:3}. 

% --------------------
\section{Experimental Setup}

\subsection{Experimental Data}
The experimental data used in this study consist of 7,446 historical ptychographic datasets acquired at the COSMIC Imaging Beamline 7.0.1.2 at the Advanced Light Source (ALS). The datasets span multiple experimental campaigns and sample types, providing diverse imaging conditions for both training and evaluation. To assess robustness and temporal generalization, the ML model is trained on a subset of historical datasets and evaluated on experimental data acquired in different years. 

\subsection{Reconstruction and Learning Parameters}
The key parameters used for both the physics-based ptychographic reconstruction and the ML model are summarized below. Unless otherwise stated, the same parameter settings are used for all experiments reported in this work.

\paragraph{CDTools Reconstruction Parameters.}
Reconstructions are performed based on the \texttt{FancyPtycho} ptychographic model class in CDTools. The automatic differentiation update for the object and the probe is conducted using the Adam optimizer with a learning rate of 0.001 and a batch size of 50. A step-based learning-rate scheduler is used to enhance convergence stability, with a step size of 50 iterations and a decay factor of 0.2. The total number of iterations is set to 100, and the ML-based fast-forward operator is applied at the beginning of iteration 5. Wavefield propagation between the probe and detector planes is modeled using a Fresnel propagator with a propagation distance of $25\,\mu\mathrm{m}$. Partial coherence is modeled using three coherent probe modes. A SHARP-style probe initialization is used to generate an initial probe estimate from the dataset by computing the mean of all measured diffraction patterns and using it as the Fourier-space probe amplitude, with all phases set to zero. The probe is constrained by a circular real-space support with a radius of 200 pixels. The reconstructed object is initialized as a complex-valued array with zero amplitude and zero-padded by 200 pixels on each side.

\paragraph{Machine Learning Model Parameters.}
The ML model employs a U-Net architecture with input and output channel width $c_{\mathrm{in}} = c_{\mathrm{out}} = 2$, base channel width $c_b = 64$, depth 4, and a channel growth rate $r = 2$. The dataset is split into training and validation subsets with a ratio of 0.98:0.02, corresponding to 7,297 datasets (58,376 patches) for training and 149 datasets (1,192 patches) for validation. Shuffling is performed at the dataset level to prevent any possible patch overlap between training and validation sets. Training is conducted for 200 epochs using a batch size of 4 and an initial learning rate of $10^{-3}$. A step-based learning-rate scheduler is applied with a step size of 100 iterations and a decay factor of 0.2. The training run requires approximately 21 hours of wall-clock time on 32 NVIDIA A100 GPUs at National Energy Research Scientific Computing Center (NERSC). %The training and validation losses are shown in Fig.~\ref{fig:2}(b)

\subsection{Baselines and Metrics}
We compare classical ptychographic algorithms and the proposed ML-based method using Poisson negative log-likelihood (NLL), defined as
\begin{equation}
\mathcal{L} = \sum_k \left( I_k - I_k^{\mathrm{exp}} \log I_k \right).
\end{equation}
where $I_k$ and $I_k^{\mathrm{exp}}$ are the predicted and measured diffraction intensities. Because ptychographic measurements follow photon-counting statistics, the Poisson NLL provides a physically appropriate data-fidelity metric, and a lower value typically means sharper features, less noise, or better-preserved structures. Importantly, since ground-truth object estimates are generally unavailable in experimental ptychography, Poisson NLL enables fair comparison across reconstruction methods using only the measured data.

% --------------------
\section{Results}
\subsection{Reconstruction Quality}
% Reconstruction quality is first evaluated on representative test samples by comparing the proposed ML-augmented approach against a standard ptychographic baseline. Two test samples are used: a nickel catalyst (test sample~1) and porous alumina particles (test sample~2). Similar agreement is observed for test sample~2, as shown in the amplitude reconstructions (g--i) and phase reconstructions (j--l). In both cases, the corresponding difference maps indicate that the ML-augmented method closely matches the baseline reconstruction while suppressing residual artifacts and noise, demonstrating high-fidelity object recovery across both amplitude and phase.

The ML-augmented ptychography approach is evaluated by comparing the reconstruction quality and converging speed against a standard ptychography baseline on 100 test samples. These test datasets were obtained from different years than the training data and span diverse sample types. The reconstruction quality is evaluated by both visual inspection and the converged Poisson NLL value. The converging speed is quantified using the iteration-to-$\varepsilon$ metric, which captures how quickly the ML-augmented method approaches the converged Poisson NLL. Specifically, iteration-to-$\varepsilon$ is defined as the first iteration at which the Poisson NLL of the ML-augmented reconstruction falls within a prescribed tolerance of the baseline reference value,

\begin{equation}
i_{\varepsilon} = \min \left\{ i : 
\left| \mathcal{L}^{(i)} - \mathcal{L}^{(i_\mathrm{ref})} \right|
\le \varepsilon \right\}.
\label{eq:i_epsilon}
\end{equation}

\begin{figure}[!ht]
    \centering
    \includegraphics[width=0.9\linewidth]{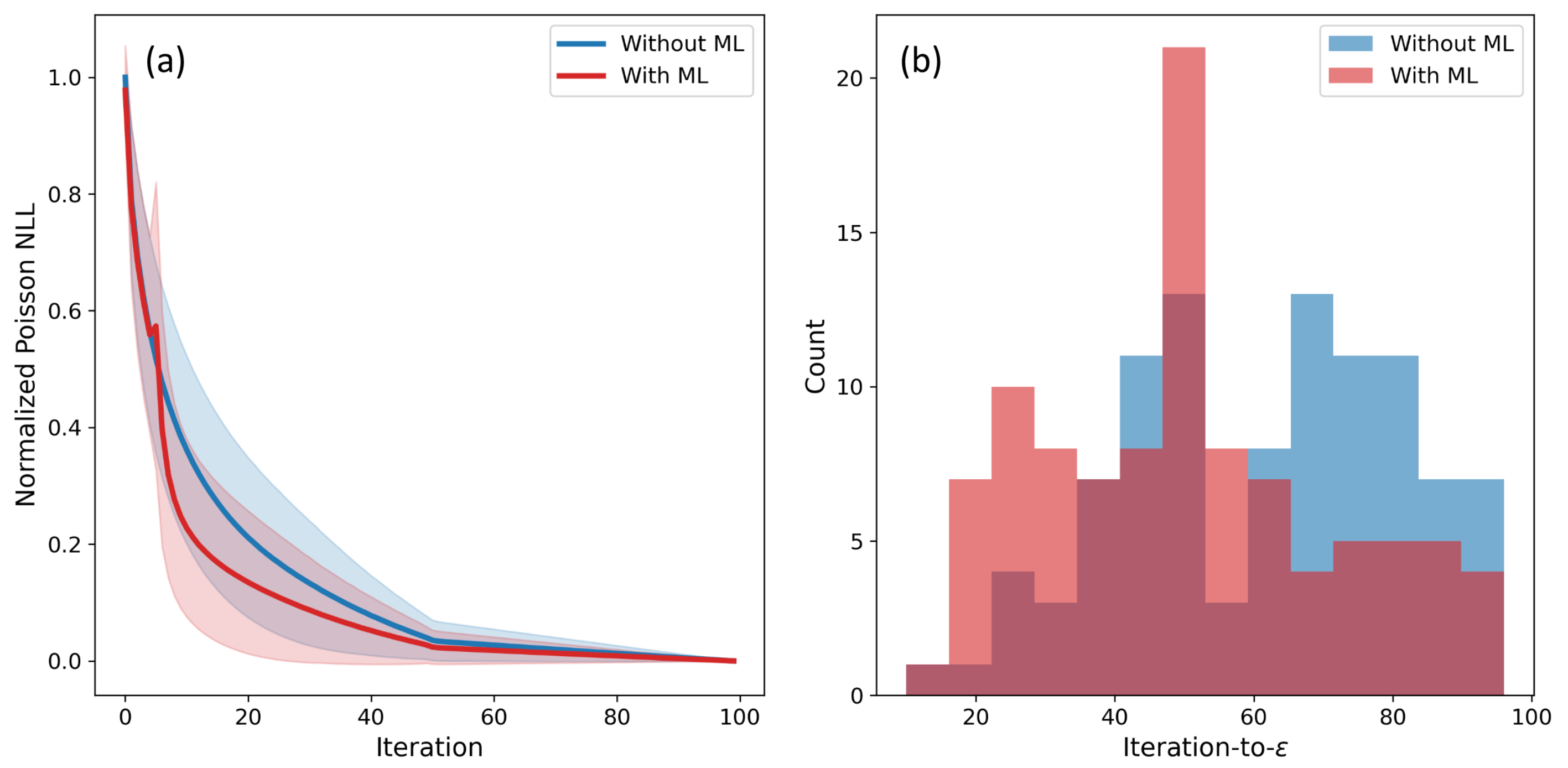}
    \caption{Statistics of reconstruction results of 100 test samples using standard and ML-augmented ptychography. (a) Averaged normalized Poisson NLL convergence curves. Shaded regions indicate ±1 standard deviation. (b) Histograms of iteration-to-$\varepsilon$.}
    \label{fig:100_samples}
\end{figure}

Here, $\mathcal{L}^{(i)}$ and $\mathcal{L}^{(i_\mathrm{ref})}$ denote the Poisson NLL of reconstruction at iteration $i$ and  the reference iteration $i_\mathrm{ref}=100$, and $\varepsilon$ is the convergence threshold, set as 2\% of the Poisson NLL range of the reconstruction. Fig.~\ref{fig:100_samples} summarizes statistically the test results of the standard and the ML-augmented reconstructions. Fig.~\ref{fig:100_samples}(a) shows the averaged Poisson NLL convergence curves. The Poisson NLL for each sample is normalized by subtracting the converged value and scaling to [0, 1], ensuring fair comparison across samples with inherently different absolute NLL magnitudes. Solid lines represent the mean and shaded regions indicate ±1 standard deviation across the 100 test samples. At the beginning, a brief increase in Poisson NLL is observed for the ML-augmented method, which coincides with the introduction of the learned prior through the fast-forward operator. After that, the Poisson NLL begins to converge faster than the standard baseline, and converges to a similar final value. The transient NLL increase reflects substantial, non-local moves through the loss landscape rather than conservative local updates. Such behavior enables the reconstruction to escape suboptimal local minima and explore more advantageous trajectories. Subsequent gradient-based ptychographic iterations re-impose consistency with the measured photon-counting statistics, guiding the reconstruction toward a statistically consistent minimum. Fig.~\ref{fig:100_samples}(b) compares the iteration-to-$\varepsilon$ histograms of the standard and the ML-augmented methods. The ML distribution is shifted clearly to the left, with the majority of samples converging between iterations 20–55, peaking around iteration 50. In contrast, the distribution of the standard baseline is spread further to the right, with a prominent cluster between iterations 60–80 and a heavier tail extending toward iteration 100. These patterns are consistent with the convergence curves in Fig.~\ref{fig:100_samples}(a).

\begin{figure}[!ht]
    \centering
    \includegraphics[width=0.95\linewidth]{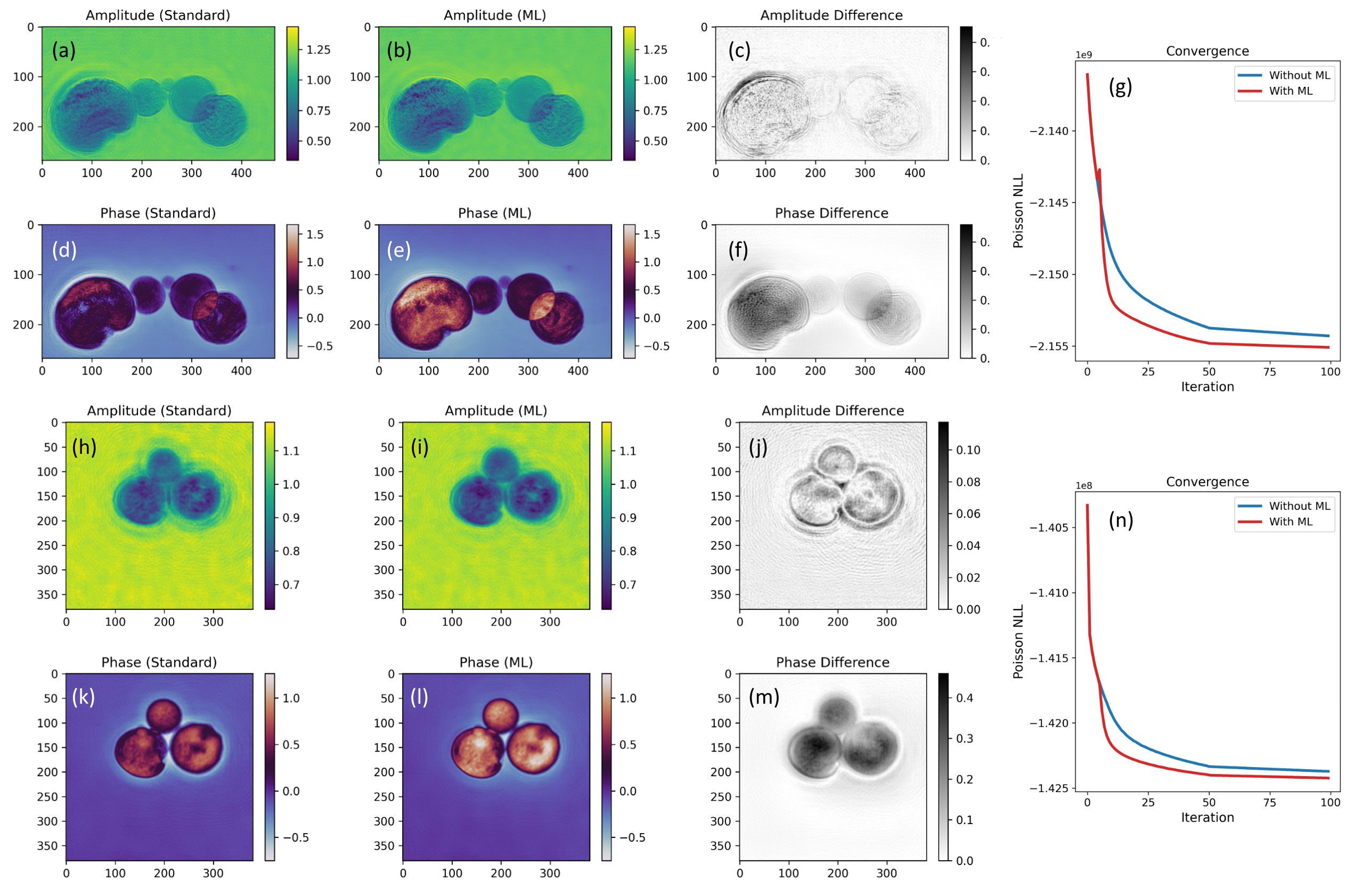}
    \caption{Comparison of standard and ML-augmented ptychographic reconstructions for two test samples (reconstructed objects). (a--c) Reconstructed object amplitude for test sample~1 obtained using the standard algorithm, the ML-augmented algorithm, and their absolute difference, respectively. (d--f) Corresponding object phase reconstructions and phase difference for test sample~1. (g) Poisson NLL curves for test sample~1. (h--j) Reconstructed object amplitude for test sample~2 obtained using the standard algorithm, the ML-augmented algorithm, and their absolute difference. (k--m) Corresponding object phase reconstructions and phase difference for test sample~2. (g) Poisson NLL curves for test sample~2. In all object reconstructions, 450 pixels are cropped from each side of the padded reconstructed object for visualization purpose.}
    \label{fig:objects}
\end{figure}

Fig.~\ref{fig:objects} provides two examples of object reconstruction via the standard and the ML-augmented ptychography. They are a nickel catalyst (test sample~1) and porous alumina particles (test sample~2). For test sample~1, the reconstructed amplitudes (a--c) and phases (d--f) produced by the two approaches are visually consistent, with the ML-augmented reconstruction yielding sharper fine structural features. Fig.~\ref{fig:objects}(g) shows the Poisson NLL curves, where the ML approach exhibits a modest increase at iteration 5, faster convergence, and a lower final Poisson NLL value compared to the standard baseline method. Similarly for test sample~2 (Figs.~\ref{fig:objects}(h--m)), the ML-augmented reconstruction yields correct structure with finer details. In the Poisson NLL results in Fig.~\ref{fig:objects}(n), the ML approach shows faster convergence and a lower final Poisson NLL value compared to the standard reconstruction baseline, but without a visible transient increase at iteration where the ML operator is applied. These two results demonstrate that the ML-augmented ptychographic method can potentially both enhance and accelerate the reconstructed objects.

\begin{figure}[!ht]
    \centering
    \includegraphics[width=0.85\linewidth]{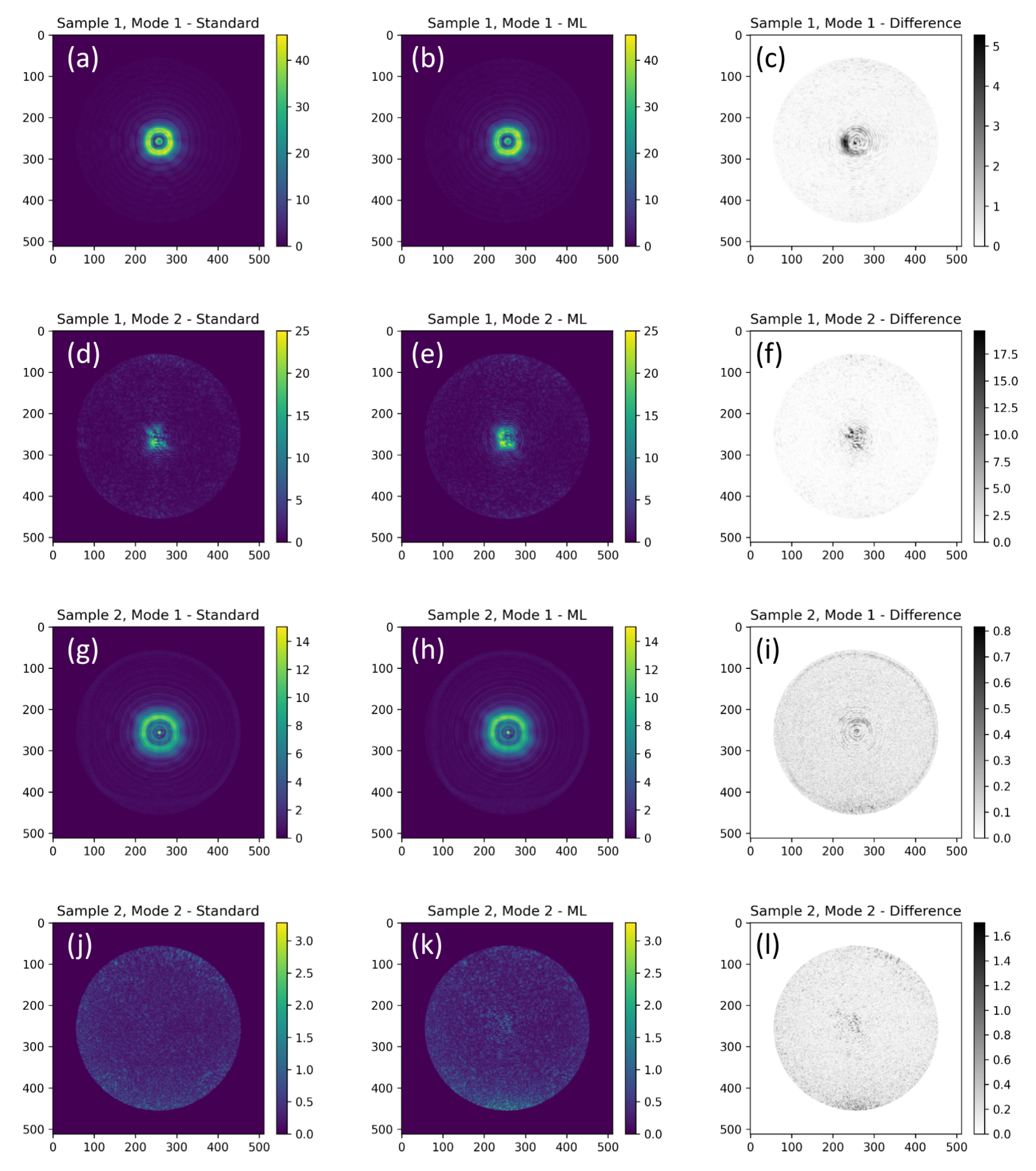}
    \caption{Comparison of standard and ML-augmented ptychographic reconstructions for two test samples (reconstructed probes). (a--c) Reconstructed probe amplitude for test sample~1, probe mode~1, obtained using the standard algorithm, the ML-augmented algorithm, and their absolute difference, respectively. (d--f) Reconstructed probe amplitude for test sample~1, probe mode~2, obtained using the standard algorithm, the ML-augmented algorithm, and their absolute difference. (g--i) Reconstructed probe amplitude for test sample~2, probe mode~1, obtained using the standard algorithm, the ML-augmented algorithm, and their absolute difference. (j--l) Reconstructed probe amplitude for test sample~2, probe mode~2, obtained using the standard algorithm, the ML-augmented algorithm, and their absolute difference.}
    \label{fig:5}
\end{figure}

Fig.~\ref{fig:5} presents a comparison of standard and ML-augmented ptychographic probe reconstructions for the same two test samples, which are reconstructed jointly with the objects. For test sample~1, probe mode~1 (a--c) and probe mode~2 (d--f) show that the ML-augmented algorithm recovers probe amplitudes that closely match those obtained with the standard reconstruction. Similar behavior is observed for test sample~2 in probe mode~1 (g--i) and probe mode~2 (j--l), where the dominant probe features are preserved. Importantly, unlike many existing ML-based ptychographic approaches that focus primarily on object recovery or assume a fixed probe, the proposed method is able to effectively recover multiple probe modes, highlighting its advantage for joint object--probe reconstruction in realistic experimental settings.

Next, the ML speedup of the two samples are examined, where we compare the wall-clock time for the standard and the ML-augmented reconstructions to reach iterations-to-$\varepsilon$. As summarized in Table~\ref{tab:convergence_speed}, the use of ML operator reduces iterations-to-$\varepsilon$ from 67 (sample 1) and 51 (sample 2) to 32 and 31, achieving a 1.83$\times$ and a 1.49$\times$ reduction in wall-clock time. These results demonstrate that ML augmentation does not compromise physical data consistency, but instead accelerates convergence while yielding reconstructions with equal or improved agreement with the measured diffraction patterns.

\begin{table}[!h]
\centering
\caption{Convergence speed comparison between the standard method and the ML-augmented method.}
\label{tab:convergence_speed}

\renewcommand{\arraystretch}{1.3}

\begin{tabular}{c|cc|cc|c}
\hline
\multirow{2}{*}{Sample}
& \multicolumn{2}{c|}{Standard} 
& \multicolumn{2}{c|}{ML} 
& \multirow{2}{*}{Speedup} \\
\cline{2-5}
& Iterations-to-$\varepsilon$ 
& Time (s) 
& Iterations-to-$\varepsilon$
& Time (s) 
& \\
\hline
1 & 67 & 74.59 & 32 & 40.69 & 1.83$\times$ \\
2 & 51 & 52.47 & 31 & 35.33 & 1.49$\times$ \\
\hline
\end{tabular}
\end{table}

\begin{figure}[!ht]
    \centering
    \includegraphics[width=0.95\linewidth]{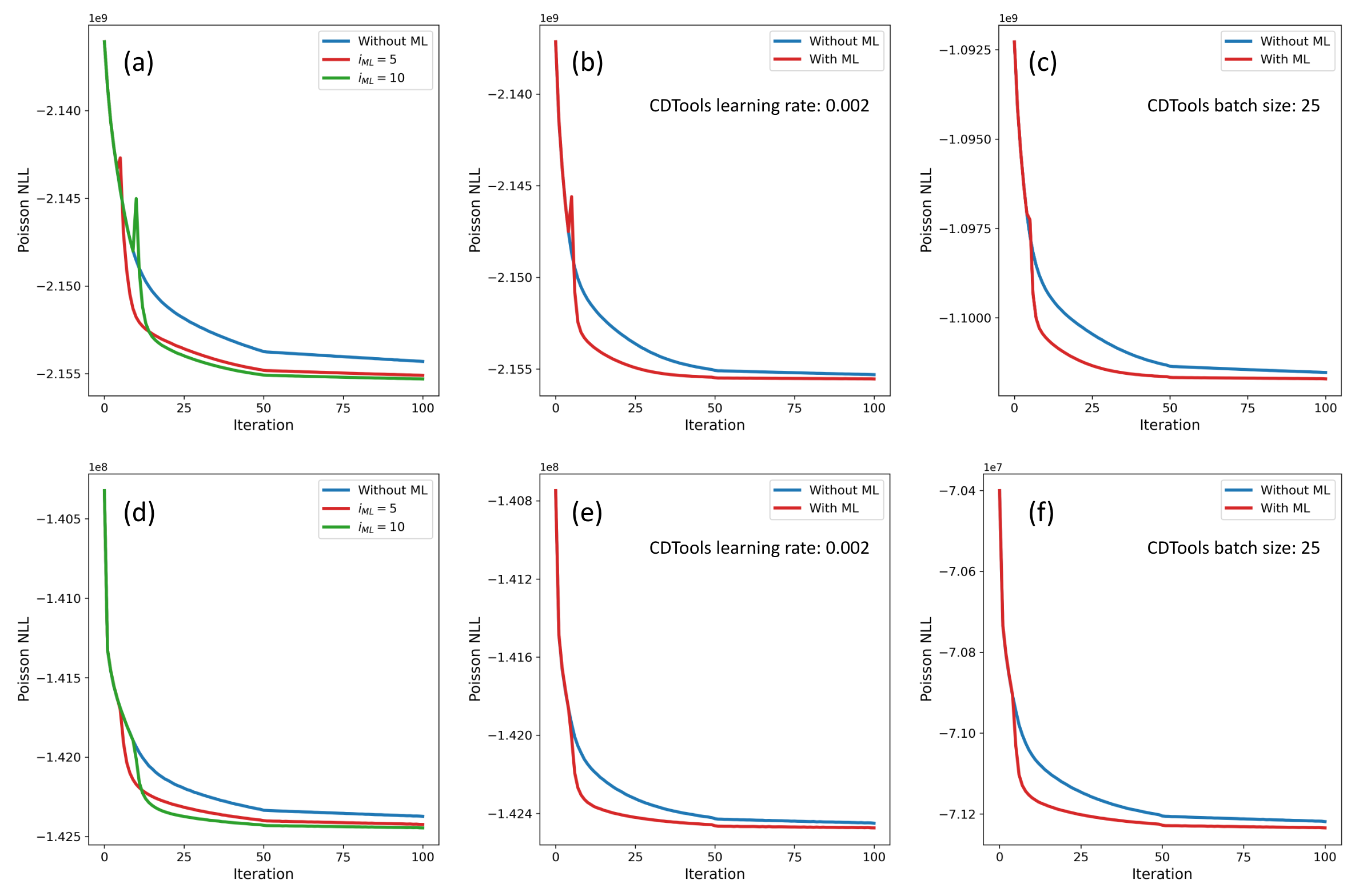}
    \caption{
    Poisson NLL curves for two test samples under different reconstruction settings. (a--c) correspond to test sample~1, and (d--f) correspond to test sample~2. (a, d) Comparison of the standard reconstruction (blue) with ML-augmented reconstructions using $i_{\mathrm{ML}} = 5$ (red) and $i_{\mathrm{ML}} = 10$ (green). (b, e) Results obtained with an increased CDTools learning rate of 0.002. (c, f) Results obtained with a smaller CDTools batch size of 25. In all cases, the ML-augmented reconstruction converges faster while reaching a final Poisson NLL lower than that of the standard algorithm, demonstrating robustness to the choice of ML insertion point and reconstruction parameters. The small "elbow" observed near iteration 50 corresponds to the activation of the scheduler for the ptychographic updates.
    }
    \label{fig:params}
\end{figure}

Fig.~\ref{fig:params} shows the Poisson NLL curves on these two test samples (Figs.~\ref{fig:params}(a--c), sample 1; Figs.~\ref{fig:params}(d--f), sample 2) under various reconstruction parameters such as $i_{\mathrm{ML}}$, CDTools learning rate and batch size. In all cases, the ML-augmented reconstruction converges more rapidly and reaches a final Poisson NLL lower than that of the standard ptychographic algorithm, indicating better statistical agreement between the predicted and measured diffraction intensities while achieving faster convergence. Figs.~\ref{fig:params}(a, d) examine the effect of the iteration at which the learned prior is introduced by comparing $i_{\mathrm{ML}} = 5$ and $i_{\mathrm{ML}} = 10$. Although the model was trained using reconstructions at iteration 5, applying it at iteration 10 still yields accelerated convergence and lowered final Poisson NLL values. This indicates that the learned operator is not narrowly tuned to a specific insertion point, but instead provides a robust update that remains effective when deployed later in the reconstruction process. Notably, for sample 1, applying ML at iteration 10 leads to more profound transient NLL increase. This may be caused by applying ML to a reconstructed object ($i=10$) that is farther away from the training data distribution ($i=5$). However, this does not necessarily lead to a less effective acceleration---applying ML at iteration 10 results in a faster convergence for both test samples. This suggests that an optimal $i_{\mathrm{ML}}$ can be explored for the best acceleration strategy. % At early iterations, a brief increase in Poisson NLL is observed for the ML-augmented method, which coincides with the introduction of the learned prior through the fast-forward operator. This increase reflects substantial, non-local moves through the loss landscape rather than conservative local updates. Such behavior enables the reconstruction to escape suboptimal local minima and explore more advantageous trajectories. Subsequent gradient-based ptychographic iterations re-impose consistency with the measured photon-counting statistics, guiding the reconstruction toward a statistically consistent minimum. 
Figs.~\ref{fig:params}(b, e) show the effect of increasing the CDTools learning rate to 0.002. Under this setting, while the overall convergence behavior remains unchanged, the transient Poisson NLL increase is more pronounced and the curves of the standard and ML-augmented reconstructions become closer, indicating reduced gain of using the ML fast-forward operator. Figs.~\ref{fig:params}(c, f) present results obtained with a smaller CDTools batch size of 25. While the transient NLL increase remains similar to the baseline case (Fig.~\ref{fig:objects}(g)), the difference between the NLL curves becomes slightly smaller for both test samples. Together, these results demonstrate that the ML-augmented reconstruction is robust across different reconstruction settings and does not depend critically on the specific parameters with which the training data were generated.

\subsection{Reconstruction Performance}
This section evaluates the computational efficiency and scalability of the reconstruction procedure relative to the standard solver. Table~\ref{tab:gpu_runtime_metrics} summarizes the scaling performance of the standard and ML methods for 100 ptychographic iterations using test sample~1. The diffraction dataset has a total size of 1.12~GB and consists of 1,125 512$\times$512 diffraction patterns. Both approaches demonstrate substantial runtime reductions as the number of GPUs increases from 1 to 4, indicating effective parallelization. The standard method achieves speedups of 1.76 and 2.70 on 2 and 4 GPUs, corresponding to parallel efficiencies of 88\% and 68\%, respectively. The ML method exhibits comparable scaling behavior, with speedups of 1.73 and 2.60 and efficiencies of 87\% and 65\% for the same configurations. While the ML approach incurs a modest runtime overhead relative to the standard method, ranging from 1.5\% on a single GPU to 4.2\% on four GPUs, this overhead remains small compared to the overall performance gains achieved through multi-GPU scaling. These results indicate that the inclusion of ML components does not significantly degrade parallel efficiency and remains compatible with efficient ptychographic reconstruction.

\begin{table}[!h]
\centering
\caption{Runtime and strong-scaling metrics for standard and ML methods over 100 ptychographic iterations}
\label{tab:gpu_runtime_metrics}

\renewcommand{\arraystretch}{1.3}

\begin{tabular}{c|cc|cc|cc|c}
\hline
\multirow{2}{*}{GPUs} 
& \multirow{2}{*}{Standard (s)}
& \multirow{2}{*}{ML (s)} 
& \multicolumn{2}{c|}{Speedup} 
& \multicolumn{2}{c|}{Efficiency} 
& \multirow{2}{*}{\shortstack{ML\\Overhead (\%)}} \\
\cline{4-7}
& & 
& Standard 
& ML 
& Standard 
& ML 
& \\
\hline
1 & 124.75 & 126.66 & 1.00 & 1.00 & 1.00 & 1.00 & 1.5 \\
2 & 71.01  & 73.19  & 1.76 & 1.73 & 0.88 & 0.87 & 3.1 \\
4 & 46.73  & 48.69  & 2.70 & 2.60 & 0.68 & 0.65 & 4.2 \\
\hline
\end{tabular}

\renewcommand{\arraystretch}{1.0}

\vspace{0.5em}
\footnotesize
\parbox{\linewidth}{
Speedup is defined as the ratio of the 1-GPU runtime to the runtime at a given GPU count for each method. Parallel efficiency is defined as the speedup divided by the number of GPUs. ML overhead is computed relative to the Standard runtime at the same GPU count.}
\end{table}

\subsection{Beamline Deployment and Integration}
The present ML-augmented method has been integrated into the CDTools ptychographic reconstruction framework and deployed for production at an operational ALS beamline. To ensure reliable operation under beamline constraints, the deployment uses a GPU-aware orchestration system based on Prefect~\cite{prefect2024}. This system automatically manages edge heterogeneous computing resources, dynamically checking GPU availability and memory to prevent reconstruction failures during high-throughput live experiments. The workflow ingests raw diffraction data and converts it into a unified intermediate format, which simultaneously feeds both the standard CDTools and ML-augmented reconstruction pipelines. Reconstruction tasks are executed within containerized environments, allowing different algorithm versions and dependencies to coexist without interference. Automatic GPU queuing and file-based GPU locking ensure exclusive access to accelerator resources while maximizing throughput across all available GPUs. 

For a ptycho-tomography dataset of 74 projections, the ML-augmented approach reduces the time-to-solution under standard beamline operating conditions. By default, a fixed 200-iteration CDTools reconstruction is used, though convergence often occurs earlier. Under this standard workflow, each projection requires $\sim$153 s, whereas the ML-augmented reconstruction converges in 50 iterations with a runtime of $\sim$41 s per projection, corresponding to an effective $\sim$4$\times$ reduction. While the speedup for a single reconstruction is 1.5--2$\times$ when comparing near-converged iteration counts, the benefit becomes substantial for complex ptychography workloads such as ptycho-tomography and X-ray magnetic circular dichroism (XMCD) ptychography, which require many reconstructions. In these cases, per-reconstruction savings accumulate, enabling low-latency feedback for live experimental steering.

% --------------------
\section{Discussion}

A key advantage of the proposed approach is that the ML fast-forward operator acts directly on the reconstructed object, rather than on diffraction data or algorithm-specific variables. As a result, it is agnostic to detector size, pixel sampling, and the specific form of the underlying iterative solver, making it naturally compatible with a plug-and-play integration into existing ptychographic reconstruction workflows. An additional strength is that the model is trained and evaluated exclusively on experimental ptychographic data, in contrast to many existing ML-based ptychographic methods that rely primarily on simulated datasets, thereby enhancing relevance to practical experimental conditions.

Critically, by preserving gradient-based optimization throughout the reconstruction process, the proposed approach ensures that final solutions reside at local minima of the experimental data fidelity term. After the ML operator advances the reconstruction state, subsequent gradient-based iterations continue to minimize the Poisson NLL with respect to the measured diffraction
patterns. This guarantees that the final reconstruction is physically consistent with experimental observations---not merely a learned approximation, but an actual optimizer of the data-fidelity objective. In contrast, purely ML-based methods that bypass iterative refinement produce proxy solutions whose correspondence to the true data fidelity minimum is neither measured nor guaranteed. Such approaches may generate visually plausible reconstructions that nonetheless fail to satisfy physical constraints or accurately represent the measured data. Physics-informed learning approaches partially mitigate this issue by embedding measurement models or physical priors into the training process, thereby encouraging consistency with underlying physics. However, even in these cases, convergence to an explicitly defined data-fidelity minimum is not typically enforced during inference. The hybrid design presented here eliminates this ambiguity: the ML acceleration step improves computational efficiency, while the preserved gradient-based framework ensures that solutions remain grounded in experimental measurements through explicit minimization of a well-defined, physics-based objective function.

While the present results demonstrate the effectiveness of the proposed approach, opportunities remain for extending its scope. A natural direction for future work is to explore training models on larger and more diverse ptychographic datasets, including data acquired at multiple light sources. Such diversity may further reduce sensitivity to instrument-specific effects and improve robustness across varying experimental conditions. An additional consideration is deployment across facilities that rely on different ptychographic reconstruction software and processing pipelines. Because the proposed ML--augmented approach is solver-agnostic, it can potentially be integrated with a range of existing iterative ptychographic algorithms. This flexibility may simplify deployment across light sources that employ different software stacks and parameter choices, facilitating broader adoption in heterogeneous experimental environments. More broadly, the proposed approach may be extensible to other computational imaging techniques, such as tomography, where reconstruction similarly relies on iterative solvers and physical forward models. In such settings, a solver-agnostic, ML--augmented framework may provide improvements in reconstruction speed while remaining compatible with existing pipelines.

% --------------------
\section{Conclusion}
In this work, we present an ML--augmented approach that accelerates iterative ptychographic reconstruction by introducing a learned fast-forward operator. Following an initial warm-up using standard iterations, the fast-forward operator advances the reconstruction toward a more converged state, after which conventional iterative updates are resumed. This approach retains the physical rigor and adaptability of established ptychographic solvers while reducing the total number of iterations needed for convergence. The model is trained on a diverse set of ptychographic datasets and evaluated on experimental data collected in a different year, demonstrating cross-year generalization to independently acquired experimental data. Relative to conventional iterative solvers, the ML-augmented method achieves comparable or improved reconstruction quality with faster convergence, as evidenced by a more rapid reduction in Poisson NLL. The method has been integrated into an existing reconstruction pipeline and deployed in production at an operational synchrotron beamline. These results highlight the promise of hybrid physics--ML approaches for accelerating ptychographic reconstructions in high-throughput, realistic experimental environments.

% --------------------
\section*{Acknowledgments}
This work was partly support from the Laboratory Directed Research and Development (LDRD) Program of Lawrence Berkeley National Laboratory under U.S. Department of Energy (DOE) Contract No. DE-AC02-05CH11231. This work was also supported by the U.S. DOE, Office of Science, Office of Basic Energy Sciences (BES), Data, Artificial Intelligence, Machine Learning at the DOE Scientific User Facilities program award ``A Collaborative Machine Learning Platform for Scientific Discovery 2.0 (MLExchange Project).'' Additional support was provided by the U.S. DOE, Office of Science, Office of Advanced Scientific Computing Research (ASCR), award ``The Transformational AI Models Consortium'' under Contract No. AC02-05CH11231. E.B, and R.K. were supported by AFOSR Grant. No. FA9550-23-1-0395. D.S. was supported by the U.S. DOE, Office of Science, BES, under Award No. DE-SC002193 and the Advanced Light Source Collaborative Postdoctoral Fellowship. This research used resources of the National Energy Research Scientific Computing Center (NERSC), a U.S. DOE Office of Science User Facility, at Lawrence Berkeley National Laboratory, from award BES-ERCAP0031444 and from ``The Transformational AI Models Consortium'' allocation under Contract No. DE-AC02-05CH11231. This research made use of resources at the Advanced Light Source (ALS), a U.S. DOE Office of Science User Facility, at Lawrence Berkeley National Laboratory, under Contract No. DE-AC02-05CH11231 from the U.S. DOE, Office of Science, BES. Use of the Advanced Photon Source (APS), a U.S. DOE Office of Science User Facility, at Argonne National Laboratory, was supported by the U.S. DOE, Office of Science, BES, under Contract No. DE-AC02-06CH11357. Use of the Linac Coherent Light Source (LCLS), a U.S. DOE Office of Science User Facility, at SLAC National Accelerator Laboratory, is supported by the U.S. DOE, Office of Science, BES, under Contract No. DE-AC02-76SF00515.

% \section*{Code Availability}
% The code used to generate the results in this study can be found at XXX.

\section*{Data Availability}
The data that support the findings of this study are available from the corresponding author upon reasonable request. 

\section*{Conflict of Interest}
The authors declare no competing interests

% --------------------

\bibliographystyle{unsrtnat}
\bibliography{references}

\end{document}